# Self-Compositional Data Augmentation for Scientific Keyphrase Generation

Maël Houbre
Nantes University, LS2N
F-44000 Nantes, France
mael.houbre@univ-nantes.fr

Florian Boudin
JFLI, CNRS, Nantes University
Tokyo, Japan
florian.boudin@univ-nantes.fr

Béatrice Daille
Nantes University, LS2N
F-44000 Nantes, France
beatrice.daille@univ-nantes.fr

Akiko Aizawa
National Institute of Informatics
Tokyo, Japan
aizawa@nii.ac.jp

## Abstract

State-of-the-art models for keyphrase generation require large amounts of training data to achieve good performance. However, obtaining keyphrase-labeled documents can be challenging and costly. To address this issue, we present a self-compositional data augmentation method. More specifically, we measure the relatedness of training documents based on their shared keyphrases, and combine similar documents to generate synthetic samples. The advantage of our method lies in its ability to create additional training samples that keep domain coherence, without relying on external data or resources. Our results on multiple datasets spanning three different domains, demonstrate that our method consistently improves keyphrase generation. A qualitative analysis of the generated keyphrases for the Computer Science domain confirms this improvement towards their representativity property.

## ACM Reference Format:

Maël Houbre, Florian Boudin, Béatrice Daille, and Akiko Aizawa. 2024. Self-Compositional Data Augmentation for Scientific Keyphrase Generation. In *Proceedings of Joint Conference on Digital Libraries (JCDL)*. ACM, New York, NY, USA, 10 pages.

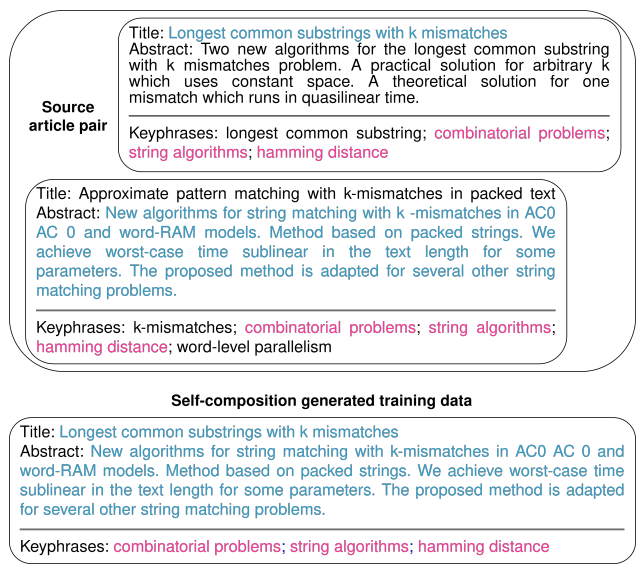

Figure 1: An illustration of a synthetic training sample generated by our self-compositional data augmentation method: the title of the first document is combined with the content of the second (in Blue) while the shared keyphrases are considered as labels (in Pink).

are useful for a variety of downstream natural language processing tasks, ranging from text summarization [36, 44] and document classification [38] to document retrieval [10, 20, 42, 43, 56]. However, for this latter task, not all keyphrases are equally important. Keyphrases that do not appear as is in the source text, refered as *absent keyphrases*, bring additional information that expands the content of the document. They thus carry added value to document retrieval systems [4, 48].

Generating keyphrases, and particularly absent ones, is a challenging task as it requires a certain level of understanding and extrapolation regarding the document. Consequently, models for keyphrase generation necessitate vast quantities of training data to be effective. In practical terms, prior studies have demonstrated that the performance of keyphrase generation improves with the size of the training dataset [18, 32, 51]. Notably, this positive impact on

## 1 INTRODUCTION

Keyphrase generation is the task of producing a set of words or phrases that highlight the key aspects of a document. Keyphrases performance is even more pronounced for absent keyphrases. However, keyphrase-labeled data is scarce and costly to produce at scale. To address this issue, data augmentation methods were introduced to expand the training data without incurring additional annotation costs. Proposed methods create synthetic training samples by either modifying existing ones through deletion, substitution or addition with external content [14, 37], or by automatically annotating unlabeled data in an self-supervised scenario [31, 51].

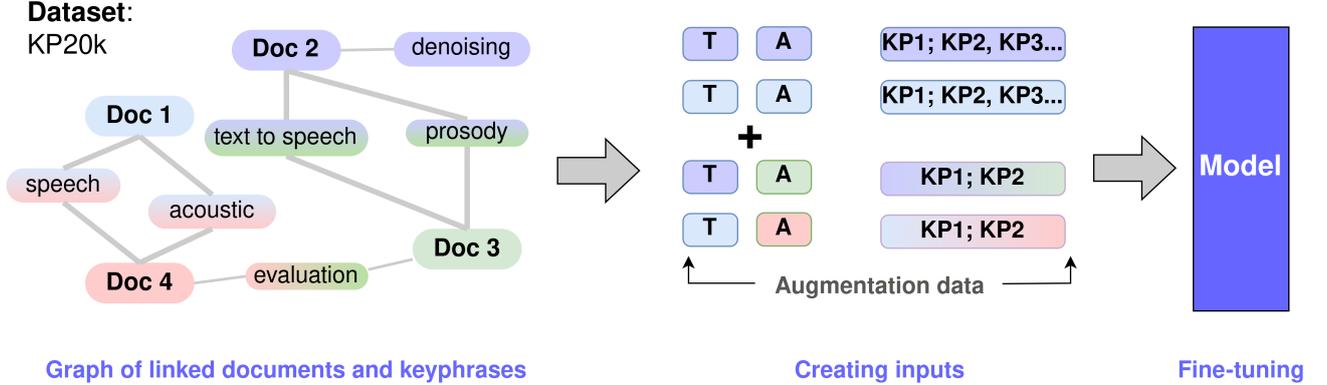

Figure 2: Overview of our augmentation approach. T, A and KP respectively stand for Title, Abstract and Keyphrase.

In this paper, we present a new data augmentation method for keyphrase generation that creates high-quality synthetic samples by combining pairs of documents from the training set. Our work is inspired by recent efforts on compositional data augmentation [2, 35], which involves creating synthetic training samples by substituting spans that occur in similar contexts. Here, we build upon this idea and propose a *self-compositional method* that generates synthetic training samples by assembling spans from similar documents, and using a combination of their keyphrases as labels. Unlike previous methods, our data augmentation does not necessitate external resources or additional unlabeled data, making it applicable to any existing dataset.

Specifically, our data augmentation method operates in two steps: first, we construct a heterogeneous graph representation of the initial training data, where documents are linked to the keyphrases assigned by their authors (§2.1). Next, we iterate over the graph to extract pairs of similar documents, determined by the number of shared keyphrases (§2.2), and assemble them to create synthetic samples (§2.3). The core idea behind our method is to leverage documents that discuss the same set of topics, here represented by their keyphrases, to create synthetic yet coherent training samples. Simply put, for each pair of similar documents, we generate a new synthetic document by combining the title of the first with the content of the second, and consider their shared keyphrases as labels (see Figure 1). We anticipate that these synthetic instances, formed by assembling the structural elements (e.g., title, abstract) of two documents connected by an indirect relation (their shared keyphrases), will be both coherent and sufficiently distant from the original data to improve the performance of keyphrase generation models.

Our contributions are as follows: 1) we introduce a *self-compositional* data augmentation method for keyphrase generation, that leverages cross-document relations to generate synthetic training samples (§2); 2) through extensive empirical experiments on several commonly-used benchmark datasets, we demonstrate that our method improves the performance of keyphrase generation models (§4); 3) we further validate the effectiveness of our method through both manual qualitative analyses of the created synthetic samples (§4.2) and of the generated keyphrases (§4.3). Furthermore, we explore its generalizability across different domains, specifically news and biomedical texts (§5).

## 2 SELF-COMPOSITIONAL DATA AUGMENTATION

Keyphrase generation involves generating a sequence of target keyphrases $y = \{y_1, y_2...y_n\}$ for a given document $X$, where each keyphrase $y_i = \{y_i^1, y_i^2...y_i^m\}$ is a sequence of words. The vast majority of training datasets for keyphrase generation, consist of bibliographic records consisting in a title, an abstract and their ground truth keyphrases, typically assigned by the authors. Therefore, the documents that we mention in this study are the concatenation of said titles and abstracts.

Given an initial dataset, our approach aims to generate additional synthetic samples to expand it, thereby enabling models to learn from more data and improve their performance. An overview of our method is presented in Figure 2. The main steps of our method are as follows: we build a graph representation of the initial dataset (§2.1), from which we extract pairs of related documents (§2.2), and then combine them to create synthetic samples (§2.3).

### 2.1 Graph Representation of the Dataset

We first build an unweighted heterogeneous graph representation $G = (E, V)$ of the initial training dataset. The set of vertices $V = (D \cup K)$ are either documents ($d_i \in D$) or keyphrases ($k_i \in K$). An edge between a document $d_i$ and a keyphrase $k_i$ indicates that $k_i$ appears in the author assigned keyphrases of $d_i$. Before building the graph, keyphrases are lowercased to enforce uniformity in the graph.

We then browse this graph to get pairs of documents that have keyphrases in common.

## 2.2 Extraction of Related Documents

Our self-compositional data augmentation method, builds upon the hypothesis that related documents –those discussing the same set of topics– are appropriate for generating synthetic training samples.

To consider two documents as related, they have to share a substantial proportion $k$ of their keyphrases. We can draw a parallel with the compositional data augmentation work from Andreas [2], where the shared keyphrases represent our similar contexts that we use to substitute spans from different documents.

As documents have a variable number of keyphrases, using a proportion for $k$ rather than a fixed number to measure similarity, prevents from forming pairs with documents that have many keyphrases, which would result in weak relations. Indeed, sharing 3 keyphrases out of 50, makes for a much weaker relation than sharing 3 keyphrases out of 5. This more restrictive criteria, improves the chances of having relevant document pairs. One example of our compositional approach is presented in Figure 1. Both articles share the keyphrases "combinatorial problems", "string algorithms" and "hamming distance". This intersection represents 60% of the longest of the two keyphrases sequences (i.e from the second document). For each document, we rank its related pairs by descending order of shared keyphrases.

Some keyphrases may appear in many documents. Therefore, some documents may be hubs in the graph (i.e they are connected to many documents) and form an important number of pairs. This could hurt the diversity of the synthetic samples. To prevent an over-representation of those documents in the synthetic samples and therefore mitigate potential over-fitting, we limit the number of pairs for a given document to $m$. This means that a document cannot form more than $m$ pairs with related documents.

## 2.3 Generation of Synthetic Samples

As illustrated in Figure 1, we form an artificial training sample from a related pair of documents by concatenating the title of the first, to the abstract of the second. As title and keyphrases have similar but complementary roles [9, 24, 28], we hypothesize that our composition on keyphrase-related document pairs, helps to ensure that the overall meaning of the pair is preserved in the artificial sample. The keyphrases intersecting in both documents are then considered as labels for the artificial sample.

## 3 EXPERIMENTAL SETTINGS

In this section, we first describe the resources and models that we have employed in our experiments as well as the training parameters. We then give details on the evaluation process and metrics.

### 3.1 Datasets

Statistics of the training and testing datasets are presented in Table 1. We applied our methodology on KP20k [33], the dataset of reference for scientific keyphrase generation. The KP20k training dataset is composed of 531k bibliographic records with a title, an abstract and author keyphrases from articles of the ACM digital library, mostly in the computer science domain. We determined the values of $k$ and $m$ empirically by trying with different values and achieved the best ratio in performance gain over dataset size with $k = 60\%$ and $m = 5$. This means that in order to keep a document pair $(D_1, D_2)$ for data augmentation, the intersection between the keyphrases of $D_1$ and $D_2$, needs to be at least 60% of the longest keyphrase sequence. This also means that we will not create more than 5 artificial examples from $D_1$. With these parameters, our method generated 149 605 artificial instances which represents a relative augmentation of 28.2%.

We test our models on four well known datasets; the test set of KP20k, SemEval2010 [23], NUS [34] and Krapivin [25]. The inputs for the models are the title and abstract of the articles.

Table 1: Statistics of the datasets used in this study.

|  | Dataset | #doc. | #kps. | KPs (%) | |
|---|---|---|---|---|---|
|  |  |  |  | Pre | Abs |
| Train | KP20k | 530.8k | 5.3 | 58.2 | 41.8 |
|  | Augmentation | 149.6k | 3.1 | 60.4 | 39.6 |
| Test | KP20k | 20k | 5.3 | 58.4 | 41.6 |
|  | NUS | 211 | 11.7 | 54.0 | 46.0 |
|  | Krapivin | 460 | 5.7 | 53.8 | 46.2 |
|  | SemEval | 100 | 14.7 | 40.1 | 59.9 |

### 3.2 Keyphrase Generation Models

We studied the impact of our method with two different models. We first chose BART [27], a generative model, pre-trained on the general domain. This model has already been studied for keyphrase generation [13, 26, 31] and has achieved state of the art performances on this task. The other one is the One2Set model [53] which is the state of the art architecture for keyphrase generation models trained from scratch.

For the fine-tuning of BART, we opted for the One2Seq generation paradigm [55] which consists in generating all the keyphrases in a single sequence, separated by a specific character or token. In our case, the keyphrases are separated by a semicolon. The keyphrases are arranged by the order achieving the best performances according to Meng et al. [32]. Present keyphrases are ordered by their order of appearance in the source text, followed by the absent keyphrases in their original order as provided by the authors. After the fine-tuning, we over-generated 20 keyphrase sequences with a beam search, to be able to get 10 unique keyphrases for our evaluations.

For One2set, we applied the normalization from the original article. The source text and the keyphrases are lowercased, the digits are replaced by a special token in the text and keyphrases. The keyphrases are in a similar order as for BART but the special token "\<peos\>" separates the absent keyphrases from the present ones. We generated the keyphrase sequences using the provided script[1].

If we did not obtain 10 unique keyphrases after the generation, we appended incorrect keyphrases until we reached that number.

---

[1] https://github.com/jiacheng-ye/kg_one2set

Table 2: Results for present and absent keyphrase generation on the test sets. ⋆ means that the results are statistically significant compared to the base model

| Model | KP20k | | | | NUS | | | | Krapivin | | | | SemEval | | | |
|---|---|---|---|---|---|---|---|---|---|---|---|---|---|---|---|---|
| | $F_1@M$ | | $F_1@10$ | | $F_1@M$ | | $F_1@10$ | | $F_1@M$ | | $F_1@10$ | | $F_1@M$ | | $F_1@10$ | |
| | Pre | Abs | Pre | Abs | Pre | Abs | Pre | Abs | Pre | Abs | Pre | Abs | Pre | Abs | Pre | Abs |
| BART | 36.5 | 2.4 | 34.6 | 4.0 | 39.4 | 3.4 | **42.8** | 5.6 | 37.6 | 3.6 | 35.3 | 5.0 | 32.6 | 1.3 | 36.4 | 3.0 |
| + augm. | **37.0**⋆ | **3.1**⋆ | **35.1**⋆ | **4.7**⋆ | **40.4** | **4.9** | 42.5 | **6.3** | **37.9** | **4.2** | **36.4** | **5.8** | **32.8** | **1.9** | **37.1** | **4.2** |
| One2Set | 36.3 | 5.4 | 36.3 | 3.5 | 37.5 | 5.8 | 37.6 | 4.4 | **35.8** | 6.3 | **35.8** | 4.0 | 31.5 | 3.4 | 31.5 | 2.7 |
| + augm. | **36.6** | **6.1**⋆ | **36.6** | **3.8**⋆ | **39.2** | **6.6** | **39.1** | **4.5** | 35.7 | **8.3**⋆ | 35.7 | **4.9** | **32.7** | **4.1** | **32.7** | **3.2** |

## 3.3 Training Parameters

We trained each model for 10 epochs and took the last checkpoint, ensuring the best performances according to Wu et al. [46]. For BART, we fine-tuned with the regular Trainer from Huggingface with a learning rate of $1e^{-4}$ and a batch size of 32. The maximum input size was set to 512 tokens and the maximum output size to 128 tokens.

For One2set, we kept the parameters from the original article which is a learning rate of $1e^{-4}$ and a batch size of 12.

## 3.4 Evaluation Metrics

For the evaluation, we compare the generated keyphrases to the author assigned keyphrases. Only exact matches are considered, which means that the model has to generate the exact keyphrase to be counted as a good prediction. In accordance with previous work, we separately evaluate present and absent keyphrase generation. The last one being more difficult, not distinguishing the different keyphrases would hurt the overall performances and give a incomplete idea of the model's capabilities. We conduct the evaluation of absent keyphrase generation in two different ways. One without distinction between the forms of absent keyphrases as well as a more precise one with the RMU characterization [4].

RMU categorization has been introduced by Boudin and Gallina [4], as absent keyphrases can take different forms. Indeed, some keyphrases may have some or even all of their constitutive words within the text and still be considered absent keyphrases. They thus present three different categories called *Reordered* (R), *Mixed* (M) and *Unseen* (U) which are defined as follows.

- **Reordered** (R): All constitutive words are in the source text but they are not contiguous.
- **Mixed** (M): Some words of the keyphrase are in the source text, others are not
- **Unseen** (U): No words of the keyphrase are in the source text

An example of an article with each kind of keyphrases is available in Table 3. According to those definitions, generating *unseen* keyphrases is therefore much more difficult than generating *reordered* keyphrases. Indeed for the latter, the model has seen all of its constitutive parts in the input and just has to copy them from the text. Whereas for *unseen* keyphrases, the model has to fully generate it. To understand better the benefits of our approach on absent keyphrase generation, we evaluate the performances on each of those keyphrase categories. For the metrics, we use F1@M and

Table 3: Example of an article having at least one keyphrase of each PRMU category. Present (P) keyphrases are in red, reordered (R) keyphrases in orange, mixed (M) keyphrases in blue and unseen (U) in black

---
**Text:** Approximate pattern matching with k-mismatches in packed text: New algorithms for string matching with k - mismatches in AC0 AC 0 and word-RAM models. Method based on packed strings. We achieve worst-case time sublinear in the text length for some parameters. The proposed method is adapted for several other string matching problems.

---
**keyphrases**: k-mismatches, combinatorial problems, string algorithms, hamming distance, word-level parallelism

---

F1@10 which are variants of the F1 measure. F1@M (respectively @10) is the F1 measure computed on the first generated sequence (respectively top 10 generated keyphrases). As the appended incorrect keyphrases would only affect the performances for unseen (U) keyphrases, we get rid of those for this finer evaluation. The F1@M metric, introduced by Yuan et al. [55] also measures the ability of the model to generate the right amount of keyphrases for an article. Before running the evaluation process, keyphrases are stemmed using the Porter Stemmer to deal with potential inflected forms. We measure the statistical significance of our results by a Student's t-test with $p < 0.05$.

## 4 RESULTS AND DISCUSSION

The results for our models (base and augmented with our method) are avalaible in Table 2 and 4. We see that for the BART model, fine-tuning on the augmented dataset significantly improves both present and absent keyphrase generation performances on KP20k. We observe the same behaviour on NUS, SemEval and Krapivin. Though the sizes of those last three datasets are not big enough to achieve statistical significance.

When it comes to One2set, the results of both models are extremely close to each other. The only statistically significant difference being a decrease of 1.4% relative (respectively 4.5% relative) on present keyphrases in F1@M (respectively F1@10). We conjecture that BART's pre-training makes it more robust to our artificial examples, whereas they may act as noise for One2Set.

Table 4: Results for absent keyphrase generation on each category of absent keyphrases on the KP20k test set. ★ means that the results are statistically significant compared to the base model

| Model | $F1@M$ | | | $F1@10$ | | |
|---|---|---|---|---|---|---|
| | R | M | U | R | M | U |
| BART | 2.8 | 1.5 | 1.7 | 5.0 | 2.9 | 2.7 |
| + augm. | 3.4★ | 2.2★ | 2.3★ | 5.7★ | 3.4★ | 3.2★ |
| One2Set | 6.6 | 3.8 | 2.4 | 6.6 | 3.9 | 2.4 |
| + augm. | 6.6 | 4.4★ | 3.1★ | 6.6 | 4.4★ | 3.1★ |

The influence of our synthetic training instances becomes clearer when we look at the performances by RMU categories in Table 4.

We notice that, the BART model trained on the augmented dataset is better on every kind of absent keyphrases. This shows that our method improves two different capabilities of the model. The improved performance in *reordered* (R) and *mixed* (M) keyphrases, suggests that the model is better at identifying the tokens from the source text that are likely to be part of a keyphrase. The improvement on *unseen* keyphrases for both BART and One2Set, suggests that our method improves the models' generative capabilities.

### 4.1 Do Our Models Generate More?

Table 5: Average percentage of generated candidates absent from the source text. ★ means that the results are statistically significant compared to the base model

| Model | BART | | One2Set | |
|---|---|---|---|---|
| | $@M$ | $@10$ | $@M$ | $@10$ |
| Base | 10.8 | 23.6 | 38.5 | 59.3 |
| + augm. | 12.0★ | 25.2★ | 40.9★ | 61.5★ |

Having better performances on absent keyphrase generation, shows that the model is better at linking a document to a concept that it did not see as is within the text. To assert if our method improves the generative capacities of a model, we compare the generation ratio of models with and without augmentation. The results on the test set of KP20k are presented in Table 5. For both architectures, training with the augmented dataset increases the ratio of absent keyphrases over present ones in the generated sequences.

Those higher generation rates, either imply that our artificial training examples have a higher proportion of absent keyphrases than the original data, or that combining a title and an abstract from related documents, reduces the title and abstract overlap, thus making the relation between those two parts of the text less explicit.

To verify those hypotheses, we first look at the proportion of absent keyphrases in the training set of KP20k and in the augmentation dataset. In Table 1, we see that the average proportion of absent keyphrases per document in the training set of KP20k, is of 41.8% when it is only of 39.6% in the augmentation dataset.

As the proportion of absent keyphrases does not seem to be the reason for an increase of abstractive capacities, we compute the percentage of elements from the title that we can retrieve in the abstract. Indeed, previous works have already considered the title as a proxy toward keyphrases [9, 21, 51], making it a central element for keyphrase generation. We make the assumption that having a title relevant to an abstract but with less intersection to it, would make the examples more abstract (i.e less explict), hence encouraging the model to generate more. In KP20k, an average 61.1% of the title can be retrieved in the abstract of each document. This value drops to 40.9% in the augmentation dataset. This confirms that the level of abstraction between title and abstract is higher in the synthetic examples.

### 4.2 Qualitative Analysis of the Augmentation Data

To ensure that our method generates good quality training examples, we conduct a manual qualitative analysis on 100 randomly picked documents and evaluate them on four aspects: The **number of keyphrases**, the **coherence between title and abstract**, the **pertinence of keyphrases**, the **proportion of generic keyphrases** (i.e keyphrases such as "analysis" that could be given to any scientific article and therefore do not bring any information about the document). The coherence between title and abstract as well as the pertinence for each keyphrases are binary. Either the title is pertinent to the abstract (respectively, a keyphrase is pertinent to the artificial training instance) or it is not.

On the number of keyphrases per document, the synthetic instances have an average of 3.1. This value is below the 5.3 of the original training data. This could influence the number of generated keyphrases by models that are trained with the One2Seq paradigm (i.e generating all the keyphrases in one sequence). Out of the 100 randomly picked examples, 8 had a title that was not relevant to the abstract. Those 8 documents had on average 27% of generic keyphrases per document. The other keyphrases were domain related but not very specific either such as "cybernetics", "online algorithms", "neural nets" or "improving classroom teaching". For the instances where the title was relevant to the abstract, the average proportion of generic keyphrases was only 13% and the average proportion of relevant keyphrases was 87%. Those results indicate that a method as simple as ours, can create good quality artificial training instances. Yet, an astonishing 24% of those examples were from duplicates. Those duplicates were not exact copies but had a copyright statement added to the original text. Also, some words had different spelling and sometimes, some of the keyphrases were changed as well (probably due to versioning). As modifying the text by adding a copyright statement or changing the spelling of some words can be considered part of data augmentation techniques, we kept those examples even if in our sample, they represented a significant part of the generated examples.

Table 6: Example of a record from KP20k's test set with authors keyphrases and generations from all models (BART, One2Set and their respective version trained on the augmented dataset)

| Text | Spoken query processing for interactive information retrieval: It has long been recognized that interactivity improves the effectiveness of Information Retrieval systems. Speech is the most natural and interactive medium of communication and recent progress in speech recognition is making it possible to build systems that interact with the user via speech. However, given the typical length of queries submitted to Information Retrieval systems, it is easy to imagine that the effects of word recognition errors in spoken queries must be severely destructive on the system's effectiveness. The experimental work reported in this paper shows that the use of classical Information Retrieval techniques for spoken query processing is robust to considerably high levels of word recognition errors, in particular for long queries. Moreover, in the case of short queries, both standard relevance feedback and pseudo relevance feedback can be effectively employed to improve the effectiveness of spoken query processing. |
|---|---|
| Authors Keyphrases | spoken query processing, information retrieval, evaluation |
| BART | spoken query processing, interactivity, speech recognition, pseudo relevance feedback |
| +augmentation | spoken query processing, information retrieval, relevance feedback, pseudo relevance feedback |
| One2set | spoken query processing, information retrieval, pseudo relevance feedback, relevance feedback, interactive information retrieval, speech analysis |
| +augmentation | information retrieval, pseudo relevance feedback, relevance feedback, speech recognition, spoken query processing, query processing, query expansion |

Table 7: Average representativity scores of the models on 20 randomly picked documents

| Model | BART | One2Set |
|---|---|---|
| Base | 62.6% | 61.4% |
| + augm. | **67.5%** | **67.0%** |

## 4.3 Manual Qualitative Analysis of the Generated Keyphrases

Previous works mostly evaluated keyphrase generation through automatic evaluation only. But generating keyphrases that do not match those assigned by the authors, does not mean that they are not relevant to the document. To see how our method influences the generation, we manually examine the generated keyphrases on 20 randomly picked documents from the KP20k test set. We examine by hand the first generated sequence for all four models (BART, One2Set and their respective versions trained with augmentation) on three criterias from Firoozeh et al. [11]. Firoozeh et al. [11] defined several linguistic, keyness and domain-specific properties to qualitatively evaluate keyphrases. Amongst those properties, we focus on the **well-formedness** of the keyphrases, their **representativity** and the **minimality of the generated sequence**.

The well-formedness consists in verifying that the generated keyphrases are well formed words or phrases. For example, "*keyphrase gen*" instead of "*keyphrase generation*" will not be considered well formed. For our evaluation, we compute the ratio of unique well formed keyphrases over the number of unique generated keyphrases (we do not consider duplicates). The minimality property measures if there are any redundancies in the generated keyphrases sequence (either complete duplicates or keyphrases referring to the same thing). We manually examine the overall keyphrase sequence. If we see duplicates or redundant keyphrases in the generated sequence, for example "keyphrase generation" and "scientific keyphrase generation", then the generated sequence is not considered minimal. The representativity property measures if the keyphrase is relevant and specific enough to the document. For example, for an article about long short term memory neural networks (LSTM), the keyphrase "neural networks" would not be representative but "recurrent neural networks" would be. For this property, we manually evaluate each keyphrase separately. To decide if a generated keyphrase is representative, we first look if we can retrieve it in the text. If yes, but it is part of a longer technical term, we check if the generated keyphrase contains most of the term's constituting words. For example, if the text contains the phrase "second-order hyperbolic differential equations", the generated keyphrase "hyperbolic differential equations" is considered representative but "second-order equations" is not. If the generated keyphrase is absent, we compare its definition to the definitions of the technical terms in the text and the definitions of the author assigned keyphrases to determine its relevance. We then calculate the ratio of the number of unique representative keyphrases over the number of unique generated keyphrases. As it appeared that both minimality and well-formedness were not affected by our approach, only representativity scores are displayed in Table 7. We note that the proportion of representative keyphrases is increased when the model is trained on the augmented dataset. A representative example of the generation behaviour of our models is available in Table 6.

## 5 GENERALIZABILITY

The results presented in the previous section, were on documents almost exclusively from the computer science domain. To evaluate the generalizability of our method, we apply it on two other domains where keyphrase generation is an active research task as well. We fine-tune BART models on those domains, as this model is the most receptive to our method (Table 2 and 4). The training parameters are the same as with BART on KP20k. We apply our data augmentation approach with the same parameters than on the computer science domain which is $k = 60\%$ and $m = 5$.

Table 8: Results on the biomedical and news domain. ★ means that the results are statistically significant, compared to the base model

|  | Model | \multicolumn{5}{c}{F1@M} | \multicolumn{5}{c}{F1@10} |
|---|---|---|---|---|---|---|---|---|---|---|---|
|  |  | Pre | Abs | R | M | U | Pre | Abs | R | M | U |
| Bio | BioBART | **36.7** | 1.1 | 0.9 | 0.8 | 1.1 | **27.4** | 2.0 | **2.1** | 1.5 | 2.1 |
|  | + augm. | 36.6 | **1.2** | **1.0** | **0.9** | **1.2** | 27.2★ | **2.1** | 2.0 | 1.5 | **2.3** |
| News | BART | **49.0** | 23.4 | 25.7 | 20.0 | 13.3 | 46.6 | **21.5** | 21.1 | 17.5 | 11.9 |
|  | + augm. | 48.7 | **23.6** | **25.9** | **20.2** | **13.7** | 47.6★ | 20.0★ | 22.1★ | 18.4★ | **12.4** |

### 5.1 Biomedical Domain

The biomedical domain is one other scientific domain where keyphrase generation is studied [15, 16]. We chose the KPBiomed dataset [18], a large scale dataset of articles from PubMed which training set has three available sizes ranging from 500k to 5.6million documents. For comparison purposes, we use the smallest size for our experiments as it has an analog size to KP20k. The augmentation approach results in 64k new training instances which is a relative augmentation of 12.8%. One hypothesis for this low number of artificial instances, is that most biomedical articles have very precise keyphrases (names of techniques, treatments or molecules etc.), resulting in a small number of articles reaching the 60% sharing threshold. As the biomedical domain may require a specific vocabulary, we opt for a domain specific BART model called BioBART [54].

The results for present and absent keyphrase generation are displayed in Table 8. We can see that the performances from the two models are very close to each other. Yet it is interesting to see that even with a limited amount of additional training documents, similar behaviour as in the computer science domain can be discerned. The performances in RMU keyphrases are sligthly improved for all categories in F1@M and the performance in *unseen* keyphrases is improved in F1@10. This implies that adding a small quantity of good quality examples can already have an effect on absent keyphrase generation.

### 5.2 General Domain

Our previous results showed that our method is relevant for two scientific domains. Yet keyphrase generation is also well developed in the general domain, particularly on news articles. We thus apply our approach on KPTimes [12], current dataset of reference for keyphrase generation on news articles. The training set of KPTimes contains 256k news articles from journals such as the New York Times with their associated keyphrases. In contrast to the previous datasets, the keyphrases for each article are not assigned by the authors but by the journals' editors.

Gallina et al. [12] showed that editors have a different annotation behaviour than authors. Keyphrases close to topic descriptors such as "*Economics*" or "*Politics and Government*" often appear in several documents, giving the dataset a much more homogeneous distribution of keyphrases amongst articles. As our data augmentation approach relies on pairs of documents sharing keyphrases, this more consistent keyphrases distribution may produce an important number of artificial training instances. Indeed, the augmentation process on KPTimes leads to 654k artificial training instances which is a relative augmentation of 252%.

The results in Table 8 show that the performances in F1@M for both models are very close to each other. However, in F1@10, he model trained on the augmented dataset has better performances in present keyphrases as well as every RMU category. For F1@10 decrease of 1.5 points in absent keyphrases but not in RMU, may be due to the bad keyphrases that are added if we do not obtain 10 unique keyphrases. As those are removed for RMU evaluation, this explains the difference of performances.

## 6 RELATED WORK

### 6.1 Keyphrase Generation Models

Meng et al. [33] was the first work that introduced keyphrase generation as it is, by training an Encoder-Decoder model for this task. The model was the first able to generate absent keyphrases. Since then, keyphrase generation has mainly been focused on two aspects: the generation paradigm and guided generation. Meng et al. [33] generated keyphrase one by one which means that a large beam search was necessary to generate different keyphrases. Then Yuan et al. [55] trained a model to generate all the keyphrases in one sequence. But as keyphrases should be an independent set and not a sequence, Ye et al. [53] developed another training paradigm, One2Set, generating a set of independent keyphrases.

Other works focused on guiding the model to generate diverse and more relevant keyphrases. Some added a correlation mechanism to the attention computation to encourage diversity [1, 7]. Others tried to guide the generation by either focusing on the title as a central element of the document [9], encoding the structure of the document [22, 30] or with topic modelling [57]. Other works focused on influencing absent keyphrase generation by first retrieving relevant keyphrases either from external sources [52] or the text itself [1, 8]. With the growing interest for pre-trained Transformer based language models on a wide range of tasks, other works applied models such as BART or T5 for keyphrase generation [13, 19, 31, 46, 47]. Those models are the current state of the art which is why this study is based on BART models.

## 6.2 Document relations for keyphrase prediction

This work is about finding related documents and harnessing those relations to create relevant artificial training examples. Several works have made use of document relations for keyphrase extraction or generation.

One popular way to use inter-document relations is through citation networks. Caragea et al. [5] improved keyphrase extraction by considering complementary features based on if a candidate keyphrase was in a citation context. In a similar manner, Gollapalli and Caragea [17] outperformed baseline PageRank based models by augmenting the representation of the content of a document with keyphrases from citation contexts. Still with citation contexts, Boudin and Aizawa [3] extracted phrases from citation contexts to get "silver labels" for unlabeled documents. Those examples with "silver labels" then served for domain adaptation training.

Another approach considered inter-document relations but with keyphrases themselves rather than contexts. Indeed, Shen et al. [41] started from the observation that an absent keyphrase for a document is very likely to be a present keyphrase in another document from the same dataset. They then created a phrase bank with candidate keyphrases present from all documents in the dataset. The best ranked candidates for each document would then be assigned to serve as additional labels with the present candidates to train a keyphrase generation model.

## 6.3 Data Augmentation for Generation Tasks

For generative tasks such as translation or summarization, data augmentation techniques are usually developed to artificially increase the size of the data in low resources scenarios, or to introduce more diversity in the data.

For translation, one common way to overcome the lack of parallel corpora is to employ back-translation on a vast amount of data in the target language or a close pivot [40, 50]. One other method is to introduce small noise in the source text to bring more diversity. Most of those works focused on modifying words of the source text by either dropping part of them [39], or replacing them with random words [45]. Others created new documents by taking elements from several inputs and concatenating them on source and target side [49]. Minus the latter, most of the presented techniques were studied for keyphrase generation [14, 32, 37]. Our work is similar to Wu et al. [49] but our concatenated elements are not randomly picked.

Data augmentation approaches for abstractive summarization, can be considered quite similar to the translation ones. For conversation summarization, Chen and Yang [6] employed similar mechanisms such as random deletion or swapping, insertion of additional text and paraphrasing. Ouyang et al. [35] extracted spans representative of a conversation structure and replaced them with others occuring in similar contexts. Loem et al. [29] also tried paraphrasing but paired with extractive summarization to create new target examples on news articles. Our work is closer to Ouyang et al. [35] as we combine elements from similar documents.

## 7 CONCLUSION AND FUTURE WORK

In this paper, we proposed a *self-compositional* data augmentation method based on keyphrases that documents share. This method does not require any additional data or resources. We showed that our method outperforms the base model in absent keyphrase generation on a wide range of datasets, while maintaining competitive performances in present keyphrases. However, One limitation of our approach, is that the heterogeneous graph which allowed us to consider the cross-document relations, was constructed with an exact matching between author assigned keyphrases. This exact matching did not allow to consider a document that has "neural network" as an author keyphrase and another that has "neural architecture" as related. Further will focus on how to better construct and use the heterogeneous graph representation to further improve the abilities of our models.

## Acknowledgements

We thank the anonymous reviewers for their valuable input on this article and our colleagues from the TALN team at LS2N for their proofreading and feedback. This work is part of the ANR DELICES project (ANR-19-CE38-0005) and was performed using HPC resources from GENCI-IDRIS (Grant 2022-[AD011013670]).